\documentclass[sigconf]{acmart}

\usepackage{geometry}
\newcommand{\bitem}{\begin{itemize}}
\newcommand{\eitem}{\end{itemize}}
\newcommand{\benum}{\begin{enumerate}}
\newcommand{\eenum}{\end{enumerate}}
\newcommand{\beq}{\begin{equation}}
\newcommand{\eeq}{\end{equation}}
\newcommand{\beqs}{\begin{equation*}}
\newcommand{\eeqs}{\end{equation*}}

\newcommand{\ep}{\varepsilon}

\newcommand{\Ds}{\mathcal{D}}
\newcommand{\tfn}{\hat \tau_{\text{fn}}}
\newcommand{\tfp}{\hat \tau_{\text{fp}}}
\newcommand{\cfp}{C_{\tfp}}
\newcommand{\cfn}{C_{\tfn}}
\newcommand{\cad}{C_{\text{ad}}}
\newcommand{\fp}{\text{fp}}
\newcommand{\fn}{\text{fn}}
\renewcommand{\qed}{\hfill\blacksquare}

\usepackage{amsmath} 
\usepackage{bm} 
\usepackage{bbm} 
\makeatletter
\@ifundefined{proposition}{%
    
}{}
\@ifundefined{definition}{%
}{}
\@ifundefined{lemma}{%
    \newtheorem{lemma}{Lemma}
}{}
\@ifundefined{corollary}{%
    \newtheorem{corollary}{Corollary}
}{}
\@ifundefined{theorem}{%
    \newtheorem{theorem}{Theorem}
}{}
\@ifundefined{corollary}{%
    \newtheorem{corollary}{Corollary}
}{}
\@ifundefined{assumption}{%
    
}{}
\@ifundefined{problem}{%
    
}{}
\@ifundefined{claim}{%
    
}{}
\makeatother

\newtheorem{innercustomgeneric}{\customgenericname}
\providecommand{\customgenericname}{}
\newcommand{\newcustomtheorem}[2]{%
  \newenvironment{#1}[1]
  {%
   \renewcommand\customgenericname{#2}%
   \renewcommand\theinnercustomgeneric{##1}%
   \innercustomgeneric
  }
  {\endinnercustomgeneric}
}
\newcustomtheorem{customclaim}{Claim}

\providecommand{\eqa}				[1]		{\begin{align}#1\end{align}}
\providecommand{\eqas}			[1]		{\begin{align*}#1\end{align*}}

\providecommand{\ie}{\emph{i.e.,}~}

\providecommand\qcomment[1]{ }

\providecommand{\realnum}					{\mathbb{R}}

\renewcommand{\(}						{\left(}
\renewcommand{\)}						{\right)}

\providecommand{\Prob}{\mathbbm{P}}

\def\Ds{\mathcal{{D}}}

\def\Xs{\mathcal{{X}}}
\def\Ys{\mathcal{{Y}}}

\usepackage{mathtools} 
\usepackage{makecell}
\usepackage{subcaption}
\usepackage{graphicx}
\usepackage{amsmath}
\usepackage{amsthm}
\usepackage{booktabs}
\usepackage{algorithm}
\usepackage{algorithmic}
\usepackage{adjustbox}

\renewcommand{\epsilon}{\ep}
\newtheorem*{remark}{Remark}
\usepackage{comment}
\usepackage{indentfirst}
\usepackage{amsmath}
\usepackage{multirow}

\AtBeginDocument{%
  \providecommand\BibTeX{{%
    \normalfont B\kern-0.5em{\scshape i\kern-0.25em b}\kern-0.8em\TeX}}}

\acmDOI{10.475/123_4}

\acmISBN{123-4567-24-567/08/06}

\copyrightyear{2022}
\acmYear{2022}
\setcopyright{acmcopyright}\acmConference[KDD '22]{Proceedings of the 28th ACM SIGKDD Conference on Knowledge Discovery and Data Mining}{August 14--18, 2022}{Washington, DC, USA}
\acmBooktitle{Proceedings of the 28th ACM SIGKDD Conference on Knowledge Discovery and Data Mining (KDD '22), August 14--18, 2022, Washington, DC, USA}
\acmPrice{15.00}
\acmDOI{10.1145/3534678.3539408}
\acmISBN{978-1-4503-9385-0/22/08}

\begin{document}

\title[PAC-Wrap]{PAC-Wrap: Semi-Supervised PAC Anomaly Detection}


\author{Shuo Li}
\authornote{The first two authors, Shuo Li and Xiayan Ji, contributed equally to this paper.}
\affiliation{%
  \institution{University of Pennsylvania}
  \city{Philadelphia}
  \state{PA} \country{US}}
\email{lishuo1@seas.upenn.edu}
\author{Xiayan Ji}
\authornotemark[1]

\affiliation{%
  \institution{University of Pennsylvania}
  \city{Philadelphia}
  \state{PA}\country{US}}
\email{xjiae@seas.upenn.edu}
\author{Edgar Dobriban}
\affiliation{%
  \institution{University of Pennsylvania}
  \city{Philadelphia}
  \state{PA}\country{US}}
\email{dobriban@wharton.upenn.edu}
\author{Oleg Sokolsky}
\affiliation{%
  \institution{University of Pennsylvania}
  \city{Philadelphia}
  \state{PA}\country{US}}
\email{sokolsky@cis.upenn.edu}
\author{Insup Lee}
\affiliation{%
  \institution{University of Pennsylvania}
  \city{Philadelphia}
  \state{PA}\country{US}}
\email{lee@cis.upenn.edu}







\begin{abstract}
Anomaly detection is essential for preventing hazardous outcomes for safety-critical applications like autonomous driving. 
  Given their safety-criticality, these applications benefit from provable bounds on various errors in anomaly detection. 
  To achieve this goal in the semi-supervised setting, we propose to provide Probably Approximately Correct (PAC) guarantees on the false negative and false positive detection rates for anomaly detection algorithms. 
  Our method (PAC-Wrap) can wrap around virtually any existing semi-supervised and unsupervised anomaly detection method, endowing it with rigorous guarantees.
  Our experiments with various anomaly detectors and datasets indicate that PAC-Wrap is broadly effective.
\end{abstract}

\ccsdesc[500]{Security and privacy~Intrusion/anomaly detection and malware mitigation}
\ccsdesc[300]{Theory of computation~Sample complexity and generalization bounds}
\ccsdesc[300]{Computing methodologies~Semi-supervised learning settings}

\keywords{Anomaly Detection, Semi-Supervised Learning, Statistical Machine Learning, PAC Learning}


\maketitle
\section{Introduction}
Anomaly detection aims to detect points that significantly deviate from the regular pattern of data and may threaten system safety.
In recent years, anomaly detectors based on machine learning algorithms have started to outperform classical methods in many tasks  \cite{10.1145/3439950, https://doi.org/10.48550/arxiv.1901.03407, Yang2022}.
 Some of these tasks are safety-critical and require rigorous guarantees on the false negative and false positive rates.
 However, machine learning-based anomaly detectors usually do not guarantee these rates by default. 

 Some methods propose using standard conformal prediction  \cite{vovk2003mondrian, balasubramanian2014conformal}, an uncertainty quantification technique, for rigorous guarantees. These methods are effective when sufficient data is given, i.e., the dataset is large enough to represent the whole data distribution. Nevertheless, we cannot make this assumption in practical settings, and hence we shall allow for some error margin incurred by the data insufficiency.  
 An alternative approach is to use training-set conditional methods, such as inductive conformal prediction  \cite{papadopoulos2008inductive}, which satisfy a Probably Approximately Correct (PAC) property \cite{valiant1984theory, vovk2012conditional, park2020pac}.
 As we will argue, this property offers more flexibility than the marginal guarantees for conformal prediction.
 Furthermore, most anomaly detection methods with rigorous guarantees only control the false positive rate (FPR). The lack of false negative rate (FNR) guarantees could limit the usefulness of a system since classifying anomalies as normal can be a consequential mistake.

\begin{figure}
    \centering
    \includegraphics[scale=0.4]{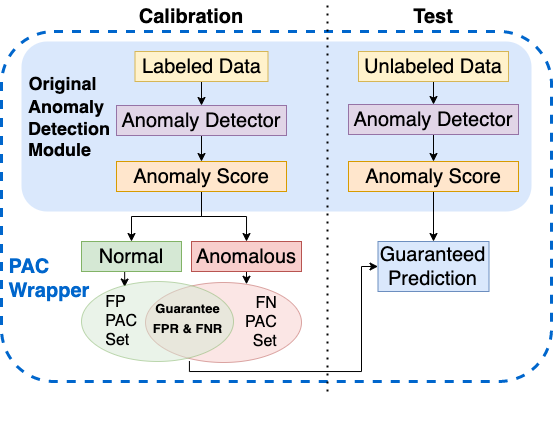}
    \vspace{-25pt}
    \caption{An overview of PAC-Wrap, which wraps around an arbitrary anomaly detector. In the calibration phase, we derive false positive (FP) PAC sets and false negative (FN) PAC sets, which guarantee the false positive (FPR) and negative rates (FNR) respectively. We take an intersection to have both guarantees. After eliminating ambiguity, it is later used at the test phase to detect anomalies with a PAC guarantee.}
    
    \label{fig:great}
    \vspace{-18pt}
\end{figure}

Hence, we propose an algorithm, named \textit{PAC-Wrap}, to add a layer ensuring a PAC guarantee on FPR and FNR to virtually any anomaly detector. In other words, PAC-Wrap acts like a wrapper that helps an anomaly detector attain a rigorous performance guarantee while keeping its internal structure intact. PAC-Wrap takes a user-specified \textit{error} level, denoted as $\epsilon$, and a user-specified \textit{confidence} level, denoted as $\delta$, to customize the guarantee.
We perform this in semi-supervised anomaly detection, where a small amount of labeled data is available  \cite{DBLP:journals/corr/abs-1906-02694, 8594877, minhas2020semisupervised}.
Our algorithm leverages the limited labeled data and provides training-set conditional guarantees, which we argue are more practical than the marginal guarantees provided by standard conformal prediction-based methods. 
Since we leverage both labeled normal and anomalous data, we can provide PAC guarantees not only on the FPR but also the FNR.

Given any trained anomaly detector that outputs the anomaly score, our wrapper method constructs false positive and false negative PAC prediction sets on the calibration datasets. These two PAC prediction sets provide PAC guarantees on FPR and FNR. Then, we propose to take the intersection of the PAC prediction sets and adopt the classification with rejection option idea \cite{herbei2006classification,bartlett2008classification}. The resulting anomaly detector guarantees FPR and FNR if it is confident about its prediction. On the other hand, if the anomaly score falls into the ambiguity region where it is not sufficiently confident about its prediction, it abstains from predicting. In cases where the rejection option is not allowed, we further propose an algorithm to eliminate the ambiguity regions. Finally, we prove that the prediction sets and the final anomaly detector are probably approximately correct. An overview of our method is in Figure \ref{fig:great}. 

We conduct experiments to validate the correctness of our theorems on both synthetic and benchmark datasets. 
Moreover, we also demonstrate that our wrapper 
can ensure that the performance of the underlying anomaly detector is rigorously guaranteed at a user-specified error and confidence level. Furthermore, to demonstrate the generalizability benefit of the PAC-based guarantee, we compare the performance of our PAC anomaly detection method to standard conformal prediction-based methods \cite{lei2014classification}. Finally, we explore the relationship between the error level, the confidence level, and the ambiguity region. 


 In summary, our contributions are as follows:
\begin{itemize}
 \item We propose to wrap PAC prediction sets around general anomaly detectors. We show rigorous guarantees on the FNR and FPR in semi-supervised anomaly detection.
\item We show that the training-set conditional PAC guarantee has both practical and theoretical benefits in generalization and flexibility compared to marginal guarantees provided by the standard conformal prediction.
\item We demonstrate empirically in simulations and on challenging benchmark datasets, using a variety of state-of-the-art anomaly detectors, that PAC-Wrap is effective.  
\item We conduct an ablation study to evaluate the tradeoff between the error level, the confidence level, and the ambiguity region. 
\end{itemize}

\vspace{-8pt}
\section{Related Work}
Conformal prediction (CP), also referred to as conformal inference~\cite{cp}, is a general approach to uncertainty quantification. It can provide finite dataset coverage guarantees under exchangeability of the data points and has been widely adopted, e.g., \cite{papadopoulos2011regression, balasubramanian2014conformal}, etc.
Closely related to our work, \cite{lei2014classification} also provides guarantees on false negative and false positive rates for classification. 
However,  \cite{lei2014classification} is based on standard conformal prediction and provides a coverage guarantee that holds marginally over the training set.
Similarly, the work \cite{guarrera2021classwise} proposes a class-wise thresholding scheme for OOD detection algorithms to maintain a comparable true positive rate across classes. Mondrian conformal prediction is a general approach to provide guarantees conditional on a general data clustering (of which class-conditional guarantees are a special case). However, their guarantees studied so far hold marginally over the training set \cite{vovk2003mondrian}.
The guarantees of conformal prediction, which hold marginally over the training set  \cite{guarrera2021classwise, lei2014classification, vovk2003mondrian} mean that the method works for most collections of training data and one test data point. 
It implies that the coverage holds for only one test data point. In contrast, the PAC guarantee we use implies that the coverage holds for most future test data points; this is more aligned with the practice setting in which a prediction method is used for many test data points.

Moving beyond standard conformal inference, \cite{chernozhukov2018exact} proposes a block permutation method to account for temporal dependence. 
EnbPI  \cite{xu2021conformal} proposes distribution-free prediction intervals for dynamic time series, 
extending CP to assume that only the residuals of a fitted model are exchangeable instead of the complete data. 
Our method differs, as we build upon the PAC framework, which---as discussed above---provides different guarantees. 
Also, our time series examples are different, as in some cases, our data points are independent time series. Thus the guarantees apply directly, similarly to previous examples such as  \cite{laxhammar:hal-01523096}.
In other cases, we take sufficiently separated subsequences of the time series that we expect them to be nearly independent, which holds for certain types of mixing conditions, as in \cite{chernozhukov2018exact}.

Inductive Conformal Prediction (ICP)  \cite{papadopoulos2002inductive}
was originally shown to have marginal guarantees, but was later shown to satisfy training-conditional, or PAC guarantees \cite{vovk2012conditional,park2020pac}. 
As discussed in \cite{vovk2012conditional}, the mathematical structure of these methods is closely related to that of tolerance regions \cite{wilks1941determination,krishnamoorthy2009statistical}. 
Inductive conformal anomaly detection \cite{papadopoulos2002inductive, kaur2022idecode} builds on ICP to guarantee a bounded false detection rate.
In different literature, there are different terminology for the two user-specified inputs. For example, the $\beta-\text{content}$ in \cite{guttman1970statistical} is equivalent to $1-\epsilon$, where $\epsilon$ is the \textit{error parameter} in \cite{10.5555/200548}. The \textit{confidence level} $\gamma$ in \cite{guttman1970statistical} is equivalent to $1-\delta$, where $\delta$ is called \textit{confidence parameter} in \cite{10.5555/200548}. In our work, we follow the terminology in \cite{10.5555/200548} and denote $\epsilon$ as the \textit{error parameter}, and $\delta$ as the \textit{confidence parameter}.
We adopt the core ideas behind this general line of work and focus on adapting it to semi-supervised anomaly detection, where both false positive and false negative rates control are essential. 

We focus on semi-supervised anomaly detection (SSAD) tasks, which have been defined in slightly different ways. 
In these definitions, given a dataset $S$, we have $m$ unlabeled data points and $n$ labeled data points, where $m \gg n$. 
SSAD definitions differ in the setup of the training and testing sets. For example, some papers  \cite{Pang2019DeepAD, DBLP:journals/corr/abs-1906-02694} assume that the training set has labeled normal and anomalous data points.
This setting is also called \textit{weakly supervised anomaly detection}. 
On the other hand,  some papers  \cite{10.1162/089976601750264965, jiang2021semi} assume that the training set only contains normal data points and the test set contains both normal and anomalous data points. We adopt the first definition of SSAD. In both our problem formulation and experimental evaluation, we assume that the training set has labeled normal and anomalous data points. 
Note that all the aforementioned semi-supervised algorithms are orthogonal to our work in that we emphasize providing a theoretical guarantee on false negative and false positive rates, whereas they focus on detector performance, like accuracy or F1-score.
\vspace{-5pt}
\section{Preliminaries}
\subsection{PAC Prediction Sets}
For independent and identically distributed (i.i.d.) training and test data, training-set conditionally valid (or, PAC) prediction sets 
\citep{vovk2012conditional,park2020pac} are guaranteed to contain the true labels for test inputs with low error level and high confidence level.
While the algorithms in  \citep{vovk2012conditional} and \citep{park2020pac} are identical, we follow the latter.
To ensure the prediction sets are small, \citep{park2020pac} solves an optimization problem to calculate the smallest prediction set (on average) while satisfying the PAC property.

Let $\Xs$ be the input space and $\Ys$ be the finite label space; let $\Ds$ denote a distribution over $\Xs \times \Ys$; let $C: \Xs \rightarrow 2^\Ys$ denote a prediction set. The probability that $C$ does not cover a test data point $(x, y) \sim \Ds$ is defined as
\begin{equation}
  L_{\Ds}(C) \coloneqq \Prob_{(x, y) \sim \Ds}[ y \notin C(x) ].  \label{pred_loss}
\end{equation}

Let $Z \sim \Ds^n$ be a held-out calibration set of i.i.d. data points from $\Ds$ with size $n$, which we can use to tune or calibrate $C$, as described below.
The goal is to find a set of a small size satisfying the PAC property, \ie given $\epsilon,\delta\in(0,1)$, 
\begin{equation*}
    \Prob_{Z \sim \Ds^n}[ L_\Ds(C) \le \epsilon ] \ge 1 - \delta,
\end{equation*} 
where the $\Prob_{Z \sim \Ds^n}$ refers to the chances of calibration succeeding.
In this case, we say $C$ is $(\epsilon, \delta)$-correct. To calculate such $(\epsilon, \delta)$-correct sets,  \cite{park2020pac} then proposes the following one-dimensional parametrization of prediction sets:
\begin{equation*}
  C_\tau(x) = \{ y \in \Ys \mid f(x, y) \ge \tau \},
\end{equation*}
where $\tau \in \realnum_{\ge 0}$ and $f:\Xs\times\Ys\to\mathbb{R}_{\ge0}$ is any given scoring function (e.g., the label probabilities output by a deep neural network).
The parameter value $\tau$ is identified by solving the following optimization problem:
\eqa{
\hat\tau = \operatorname*{\arg\max}_{\tau \in \realnum_{\ge 0}}~ \tau ~~ \text{subj. to} \sum_{(x, y) \in Z} \mathbbm{1}\( y \notin C_\tau(x) \) \le k^*,
\label{eqn:algorithm}
}
where
\eqas{
k^* =\operatorname*{\arg\max}_{k\in\mathbb{N}\cup\{0\}}~k\qquad\text{subj. to}\qquad F(k;n,\epsilon) \le \delta,
}
where $F(k; n, \epsilon)$ is the cumulative distribution function of the binomial random variable $\text{Binomial}(n, \epsilon)$
with $n$ trials and success probability $\epsilon$. 
Maximizing $\tau$ corresponds to minimizing the prediction set size. 
This is equivalent to inductive conformal prediction with the non-conformity measure $f(x,y)$, as explained in \cite{vovk2012conditional}.
Lastly, we have the following theorem:
\begin{theorem}[\citep{vovk2012conditional,park2020pac}] \label{thm:pred_set}
$C_{\hat\tau}$ is $(\epsilon, \delta)$-correct for $\hat\tau$ as in \eqref{eqn:algorithm}.
\end{theorem}

\begin{remark}
The optimization  problem \eqref{eqn:algorithm} returns the trivial solution $\hat\tau=0$ if the the optimization problem is infeasible.
\end{remark}

\subsection{Semi-supervised Anomaly Detection}

We assume each labeled data point consists of features and a label, $z_i = (x_i, y_i)$, with $y_i = 1$ indicating an anomaly (positive) and $y_i=0$ indicating a normal (negative) data point.  In a general semi-supervised anomaly detection setting, given an observed labeled data set $\{z_1, \ldots, z_N, z_{N+1}, \ldots, z_{N+K}\}$, we assume that $\{z_{N+1}, \ldots, z_{N+K}\}$ with $K \ll N$ is a small set of anomalies. At the same time, the rest of the data points are normal. We then use the observed data set as the calibration set, which contains both normal and anomalous data points. Finally, after getting the trained anomaly detector from the original semi-supervised training procedure, we calculate the PAC thresholds on the calibration set to identify anomalies.
\vspace{-5pt}
\section{Method}

Suppose we are given a semi-supervised anomaly detector $d: \Xs \to \mathbb{R}$ which maps input $x \in \Xs$ to an anomaly score.
We construct PAC prediction sets wrapped around $d(x)$ to control both false positive rate (FPR) and false negative rate (FNR). With our previous definition of positives, FPR is the rate of falsely classifying the normal class as anomalous:
\begin{equation*}
    FPR = \Prob(\hat{y} = 1 \mid y=0).
\end{equation*}
Further, FNR is the rate of erroneously predicting the anomalous class as normal:
\begin{equation*}
    FNR = \Prob(\hat{y} = 0 \mid y=1).
\end{equation*}
The control of the two rates is accomplished by replacing the original prediction error loss (as in \eqref{pred_loss}) with one that considers either FNR or FPR, which we use to construct a false negative PAC prediction set and a false positive PAC prediction set.
We then propose to take the intersection of the two sets to provide a combined guarantee, which inevitably introduces ambiguity regions. 
Lastly, we propose a strategy to remove such ambiguity by considering the relative position of the two prediction sets. 

\subsection{Conditional Prediction Sets}
In this section, we illustrate in detail our pipeline of loss modification, threshold derivation and the PAC prediction sets construction.

\textbf{False positive PAC prediction set.} 
Let the false positive PAC prediction set be $\cfp$. The loss of $\cfp$ is calculated on the normal data distribution $\Ds_{\text{nm}}$, and it is defined as:
\begin{equation}
	L_{\Ds_{\text{nm}}}(\cfp) = \mathbb{E}_{(x,y)\sim \Ds_{\text{nm}}} \ell^{01}_{\fp}(\cfp, x, y),
	\label{eqn:cfp}    
\end{equation}
where $\mathbb{E}_{(x,y)\sim \Ds_{\text{nm}}}(\cdot)$ means taking the expectation over the normal data distribution, and
$\ell_{\fp}^{01}(\cdot) \coloneqq \mathbbm{1}( y \notin C_{\hat\tau_{\fp}}(x) )$. In other words, $\ell^{01}_{\fp}(\cfp, x, y)$ indicates whether the correct label $0$ is not included in $\cfp(x)$. 

Let $Z_{\text{nm}} \sim D_{\text{nm}}^n$ be an independent calibration set of i.i.d. data points from $D_{\text{nm}}$. Given a user-specified $(\epsilon_\fp, \delta_\fp)$, 
we construct $\cfp$ by identifying the optimal $\hat \tau$ in equation \eqref{eqn:algorithm} via binary search using $Z_{\text{nm}}$. We denote the identified $\hat \tau$ as $\tfp$. 
Then, we construct the $\cfp(x)$ for $\hat y$ based on $d(x)$ in the following way:
\begin{equation}
    \cfp(x) \coloneqq \begin{cases}
    \{1\},& \text{if } d(x) \geq \tfp\\
    \{0,1\},              & \text{otherwise}
\end{cases}.
\label{fp_construct}
\end{equation} In other words, we predict the set \{1\} for $x$ with anomaly scores above $\tfp$, and \{0, 1\} otherwise. We have Corollary \ref{th_cfp} on the false positive PAC prediction. See Appendix \ref{pf_cfp} for a proof.
\begin{corollary}
$\cfp$ is $(\epsilon_\fp, \delta_\fp)$-correct for $\tfp$ identified from \eqref{eqn:algorithm} using loss function \eqref{eqn:cfp}.
\label{th_cfp}
\end{corollary} 

 Given an input $x$ and $\cfp$, we can make a class label prediction as:
\begin{equation}
\label{det_fp}
    \hat y_\fp = \mathbbm{1}(0  \notin \cfp(x)).
\end{equation}
In other words, we identify the current data point as anomalous if label 0 is not included in $\cfp$, and as normal otherwise. Then, we have Theorem \ref{th_fp} on the false positive PAC prediction set. See Appendix \ref{pf_fp} for a proof.

\begin{theorem}
$\cfp$ provides a PAC guarantee on the false positive rate:
\begin{equation*}
    \Prob_{Z_{\text{nm}} \sim D_{\text{nm}}^n} \left[ \Prob_{(x,y) \sim D_\text{nm}}(\hat y_\fp = 1 \mid y=0) \le \epsilon_\text{fp} \right] \ge 1 - \delta_\text{fp}.
\end{equation*}
\label{th_fp}
\end{theorem}

\textbf{False negative PAC prediction set.} 
We denote the false negative PAC set by $\cfn$. The loss of $\cfn$ is calculated on the anomalous data distribution $\Ds_{\text{ano}}$, and it is defined as: 
\begin{equation}
    	L_{\Ds_{\text{ano}}}(\cfn) = \mathbb{E}_{(x,y)\sim D_\text{ano}} \ell^{01}_{\fn}(\cfn, x, y),
	\label{eqn:cfn}
\end{equation}
where $\ell_{\fn}^{01}(\cdot) \coloneqq \mathbbm{1}( y \notin C_{\hat\tau_{\fn}}(x) )$.  In other words, $\ell^{01}_{\fn}(\cfn, x, y)$ indicates whether the correct label $1$ is not included in $\cfn(\cdot)$.

Let $Z_{\text{ano}} \sim D_{\text{ano}}^n$ be an independent calibration set of i.i.d. data points from $D_{\text{ano}}$. Given a user-specified $(\epsilon_\fn, \delta_\fn)$, 
 we construct $\cfn$ by identifying the optimal $\hat \tau$ in equation \eqref{eqn:algorithm} via binary search using $Z_{\text{ano}}$. We denote the identified $\hat \tau$ as $\tfn$. We then construct the false negative PAC prediction set by
\begin{equation}
C_{\hat\tau_{f
n}}(x) \coloneqq \begin{cases}
    \{0,1\},& \text{if } d(x) \geq \tfn\\
    \{0\},              & \text{otherwise}
\end{cases}.
    \label{fn_construct}
\end{equation}

Then, we have the following Corollary for the false negative PAC prediction set. See Appendix \ref{pf_cfn} for a proof.
\begin{corollary}
$\cfn$ is $(\epsilon_\fn, \delta_\fn)$-correct for $\tfn$ identified from \eqref{eqn:algorithm} using loss function \eqref{eqn:cfn}.
\label{th_cfn}
\end{corollary} 


Moreover, similarly to above, we can define 
\begin{equation}
    \label{det_fn}
    \hat y_\fn = \mathbbm{1}(1 \in \cfn(x)).
\end{equation}
Finally, we have the associated PAC guarantee for the false negative prediction set in Theorem \ref{th_fn}. See Appendix \ref{pf_fn} for a proof.
\begin{theorem}
$\cfn$ provides a PAC guarantee on false negative rate:
\begin{equation*}
    \Prob_{Z_{\text{ano}} \sim D_{\text{ano}}^n} \left[
    \Prob_{(x,y) \sim D_\text{ano}}(\hat y_\fn = 0 \mid y=1)\le \epsilon_\fn \right] \ge 1 - \delta_\fn.
\end{equation*}
\label{th_fn}
\end{theorem}
\vspace{-20pt}
\subsection{Anomaly detection with ambiguity region}
\label{subsec:adar}
We aim to use both false positive and false negative PAC prediction sets so that both rates are controlled at the same time. 
Consequently, we propose to combine false positive and false negative PAC prediction sets via taking their intersection:
\begin{equation*}
\cad (x) \coloneqq C_{\tfn}(x) \cap C_{\tfp}(x).
\label{comp_construct}
\end{equation*}
There are four possible values of the intersection, depending on the relative position of the anomaly score $d(x), \tfn$ and $\tfp$, listed in Table \ref{tab: intersection}.
\vspace*{-5pt}
\begin{table}[H]
\begin{tabular}{c|c|c}
\hline
 $d(x)$ & $< \tfn$ & $\geq \tfn$          \\ \hline
$<\tfp$ & 0 &    $\{0, 1\}$       \\ \hline
$\geq \tfp$ & $\emptyset$ &     1      \\ 
\hline
\end{tabular}
\caption{The four possible values for $\cad(x)$. }
\label{tab: intersection}
\end{table}
\vspace*{-20pt}

\begin{figure}
    \centering
    \subfloat[No Overlap]{{\includegraphics[width=0.43\linewidth]{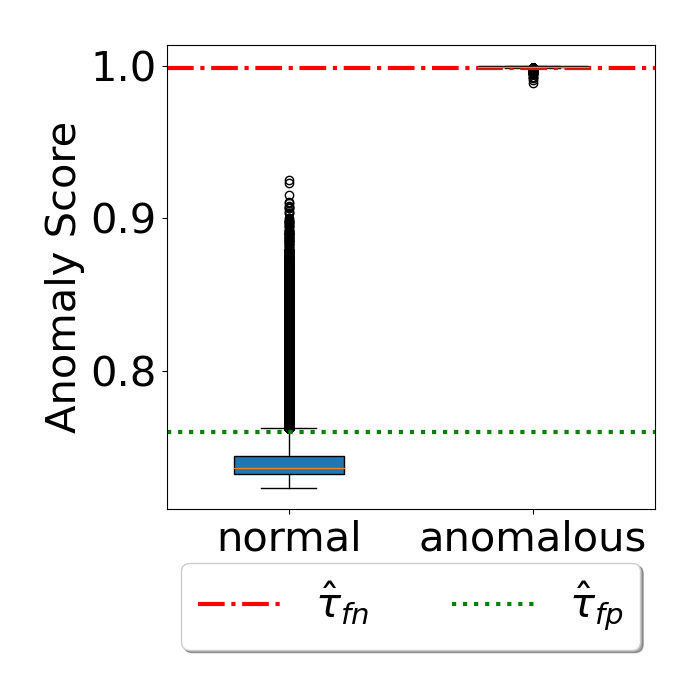} } \label{gd_a}}
    \qquad
    \subfloat[Small Overlap]{{\includegraphics[width=0.43\linewidth]{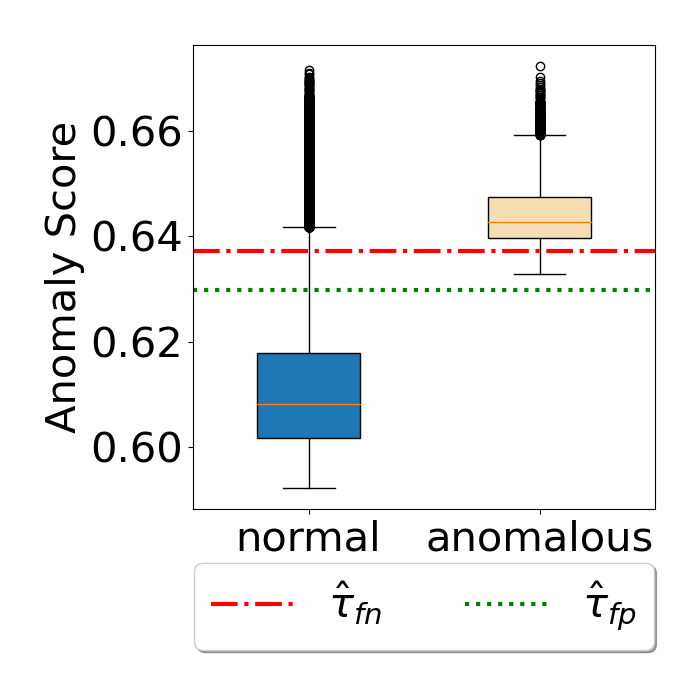} } \label{gd_b}}
    \qquad
    \subfloat[Large Overlap]{{\includegraphics[width=0.43\linewidth]{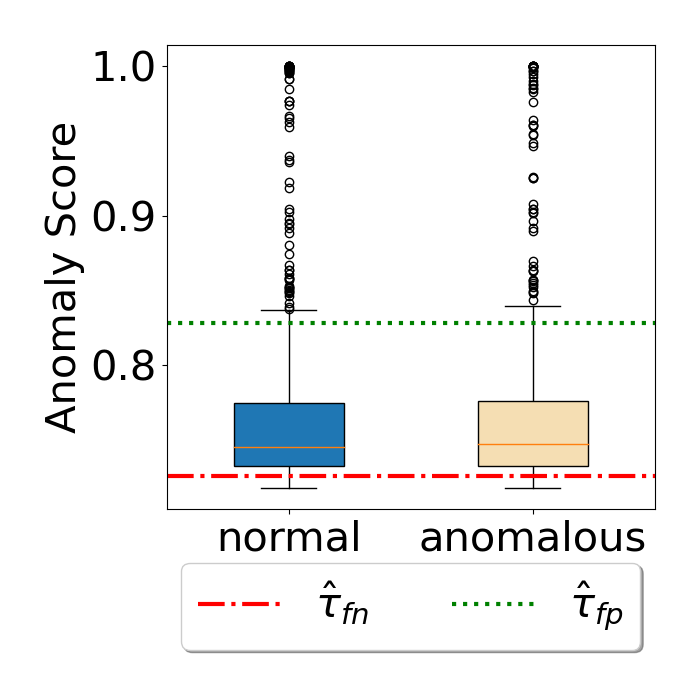} } \label{bd_a}}
    \vspace{-8pt}
    \caption{An illustration of ambiguity region cases. $\tfn \geq \tfp$ happens when the overlap between the normal and anomalous class is zero or small, as in \ref{gd_a} and \ref{gd_b}; otherwise, $\tfp \geq \tfn$ happens, as in \ref{bd_a}.
    }
    \label{fig:cases}
    \vspace{-13pt}
\end{figure}

Given an input $x$, if its anomaly score $d(x)$ falls into the interval $[\tfp, \tfn]$, (or $[\tfn, \tfp]$), $\cad$ will contain zero or two labels, which is ambiguous. 
Therefore, the interval $[\tfp, \tfn]$ (or $[\tfn, \tfp]$) is defined as the \textit{ambiguity region}, denoted as $\mathcal{U}$. Lemma \ref{ambiguity} further explains the setting when $\mathcal{U}$ occurs. 
See Appendix \ref{pf_lma} for a proof. Intuitively, when there is zero or a small overlap between the normal and anomalous classes, the ambiguity region $\mathcal{U}$ is $[\tfp, \tfn]$. This corresponds to the case $\cad=\emptyset$. 
A visualization of this case is in Figure \ref{gd_a} and Figure \ref{gd_b}, where there is no overlap (\ref{gd_a}) or little overlap (\ref{gd_b}) between the normal and anomalous classes.


\begin{lemma}
Let $k^*_\fp$ and $k^*_\fn$ be the solutions of \eqref{eqn:algorithm} when identifying false positive and false negative PAC prediction sets respectively. We have that
\begin{equation}
    \begin{split}
        & \tfn \geq \tfp  \\ 
        &\iff \\ &\sum_{(x,y) \in Z_{\text{nm}}} \mathbbm{1}(d(x) > \tfn) < k^*_\fp \\ \text{and} & \sum_{(x,y) \in Z_{\text{ano}}} \mathbbm{1}(d(x) < \tfp) < k^*_\fn.    
    \end{split}
    \label{condition}
\end{equation}
\label{ambiguity}
\end{lemma}

In this case, we predict a class label as
\begin{equation}
    \hat y_{\text{ad}} \coloneqq \begin{cases}1, & \cad(x) = \{1\}\\
    0, & \cad(x) = \{0\}\\
    *, & \cad(x) = \emptyset 
\end{cases},
\label{y_hat}
\end{equation} where $*$ means \textit{abstaining from predicting}. This classification with rejection option idea is similar to \cite{herbei2006classification,bartlett2008classification}, where a classifier could abstain from classifying an input if the classifier is not sufficiently confident about its prediction.


Further, we define the error rate $\text{ERR}(\Ds)$ over the distribution $\Ds$  as the probability that the prediction is not equal to the label:
\begin{equation*}
    \text{ERR}(\Ds) = \mathbb{P}_{(x,y) \sim \Ds}( y \neq \hat y|x ).
\end{equation*}
    Then, we have the following theorem  about the resulting anomaly detector. See Appendix \ref{pf_ad} for a proof.
    
\begin{theorem}
If $\cfn$ is $(\epsilon_\fn, \delta_\fn)$-correct, $\cfp$ is $(\epsilon_\fp, \delta_\fp)$-correct, and $\tfn \geq \tfp$, with probability at least $1-\delta_{\text{ad}}$, where $\delta_{\text{ad}}=\delta_\text{fn}+ \delta_\text{fp}$, the error rate $\text{ERR}(\Ds)$ is no greater than $\epsilon_{\text{ad}}$, where $\epsilon_{\text{ad}} = \max{(\epsilon_\text{fp}, \epsilon_\text{fn})}$, i.e.,
$$
\Prob_{Z \sim \Ds_{n}}\left[ \text{ERR}(\Ds) \le \epsilon_\text{ad}) \right] \ge 1 - \delta_{\text{ad}}.
$$
\label{th_cad}
\end{theorem} 
\vspace*{-15pt}
In contrast, $\cad = \{0,1\}$, i.e.,  $\mathcal{U}=[\tfn, \tfp]$, happens when the overlap between normal and anomalous class is large, see Figure \ref{bd_a}. According to Lemma \ref{ambiguity}, this arises when condition in \eqref{condition} fails.
In this case, the anomaly detector cannot distinguish anomalies from normal data points and therefore cannot satisfy the false positive and false negative constraints at the same time. 

Thus, we have to either find a better anomaly detector or relax the constraint for the error or confidence. 
We propose Algorithm \ref{Strategy} to relax the error constraint. Intuitively, this algorithm first checks whether $\tfn > \tfp$ or not. If not, this algorithm increases the $\epsilon$ and recalculates $\tfn$ and $\tfp$. After $\tfn > \tfp$ is satisfied, this algorithm returns the resulting $\tfn', \tfp'$, $\epsilon$ and $\delta$. Here, we use the linear search strategy where we increase the $\epsilon$ by (say) $0.1$ at each iteration. Alternatively, we could also double $\epsilon$ at each iteration, which may be faster to find a feasible $\epsilon$, but the result may be looser. 
The confidence constraint can be relaxed similarly, but the effect is less salient than that of the error constraint.


\vspace{-8pt}
\begin{algorithm}
\begin{algorithmic} 
\STATE \textbf{Input:} {$\tfn, \tfp, \epsilon_{\text{fn}},\epsilon_{\text{fp}}, \delta_{\text{fn}}, \delta_{\text{fp}}$, $\Delta$ (default $\Delta = 0.1$).}
\STATE \textbf{Output:} {$\tfn', \tfp', \epsilon, \delta$.}
\WHILE{$\tfn < \tfp$ and $\epsilon_{\text{fn}}$,  $\epsilon_{\text{fp}} \leq 1$}
\STATE $\epsilon_{\text{fn}}$,  $\epsilon_{\text{fp}}= \epsilon_{\text{fn}}+\Delta$, $\epsilon_{\text{fp}}+\Delta$.

\STATE Re-calculate $\tfn, \tfp$ using equation \eqref{eqn:algorithm} with $\delta_{\text{fn}}, \delta_{\text{fp}}$ correspondingly.
\ENDWHILE
\STATE $\epsilon, \delta = \max(\epsilon_{\text{fn}}, \epsilon_{\text{fp}}), \delta_{\text{fn}}+\delta_{\text{fp}}$.
\STATE $\tfn', \tfp' = \tfn, \tfp$.
\STATE Return $\tfn', \tfp', \epsilon, \delta$.
\end{algorithmic}
\caption{Relaxing the error constraint}
\label{Strategy}

\end{algorithm}
\vspace{-1pt}
\subsection{Anomaly detection with certain prediction}
If one is not allowed to abstain from making a prediction, the ambiguity region $\mathcal{U}$ must be removed. In this case, after satisfying $\tfn' \geq \tfp'$, we could pick an arbitrary threshold $\tau \in [\tfp', \tfn']$, e.g., $\tau = (\tfn'+\tfp')/2$, and the guarantees will still hold. We state this claim in Theorem \ref{th_final}. See Appendix \ref{pf_th_margin} for a proof.



\begin{theorem}
After using Algorithm \ref{Strategy} and picking an arbitrary threshold $\tau \in [\tfp', \tfn']$ for an anomaly detector, its error rate is at most $\epsilon$, with probability at least $1-\delta$.
\label{th_final}
\end{theorem}

\section{Experimental results}
We apply PAC-Wrap to several anomaly detectors, and on both i.i.d. and time series anomaly detection datasets. The experiments support that PAC-Wrap enables PAC guarantees on the false positive rate (FPR) and  false negative rate (FNR). In addition, we compare with standard class-conditional conformal prediction \citep{lei2014classification}. These experiments empirically support that PAC-Wrap performs well in a variety of scenarios, and compares favorably to standard conformal prediction-based methods.

We address several questions to demonstrate the effectiveness of PAC-Wrap:
\begin{itemize}
    \item \textbf{Q1 (Empirical Validation)}: Are Theorem \ref{th_fp} and \ref{th_fn}, empirically supported by results on both synthetic and benchmark datasets?
    \item \textbf{Q2 (Wrapper Effect)}: How does our wrapper affect the underlying anomaly detector's error rates?
    \item \textbf{Q3 (Baseline Comparison)}: How does our work compare to standard class-conditional conformal prediction methods?
    \item \textbf{Q4 (Ablation Study)}: How do different combinations of $\epsilon$ and $\delta$ affect the ambiguity region?
\end{itemize}

\subsection{Datasets, Anomamly Detectors, and Metrics}
\subsubsection{Datasets}
\label{d}
We first describe the synthetic and benchmark datasets used in \textbf{Q1}.
We generate a synthetic dataset by sampling i.i.d. normal and anomalous data points from two clusters, each normally distributed in 6-dimensional space with the same covariance matrix but with different means
$\mu_{\text{normal}},  \mu_{\text{anomalous}}\in \mathbb{R}^6$, which are selected so that the two classes are separated by a margin of 5.
Let $I_p$ be the $p$-dimensional identity matrix with $p=6$,
and $\sigma^2$ be a uniformly random value drawn over $[1, 100]$. 
We have:
\begin{align*}
    X_{\text{normal}} &\sim \mathcal{N}(\mu_{\text{normal}}, \sigma^2 I_p) \\
    X_{\text{anomalous}} &\sim \mathcal{N}(\mu_{\text{anomalous}}, \sigma^2 I_p).
\end{align*}
To simulate the semi-supervised problem, we generate $100,000$ normal data points as the training set, another $2,000$ normal and $2,000$ anomalous data points as the calibration set, and finally $50,000$ normal and $50,000$ anomalous data points as the test set.
The benchmark dataset \textit{thyroid} is a UCI Machine Learning Repository \cite{Dua:2019} dataset that contains around 7,200 data points. It treats the \textit{hypothyroid} disease as an anomaly. We randomly sample 80\% of the normal data points from the \textit{thyroid} dataset to form the training set. We then take the remaining 20\% of the normal data and the anomalous data points to form the calibration and the test set, with the calibration set taking up 30\% and the test set taking up 70\%. 

In \textbf{Q2}, we experiment on the benchmark semi-supervised anomaly detection datasets \textit{campaign}, \textit{celeba}, and \textit{census} that are also used in \citep{Pang2019DeepAD}. The \textit{campaign} dataset contains direct marketing campaigns (phone calls) and asks to predict whether a given client will subscribe to a term deposit. Successful campaigning records account for approximately 10\% records and are regarded as anomalies. The \textit{celeba} dataset is an image dataset of more than 200K celebrity images. In this task, the anomaly detector detects bald celebrities as anomalies, which account for less than 3\% of celebrities. The \textit{census} dataset is extracted from the US census bureau database and aims to detect the high-income people that comprise about 6\% of the data as "anomalies". In contrast to the typical supervised classification setting, these datasets are highly imbalanced. In other words, only a small portion of labeled data points are anomalous.
 


We also conduct experiments on two time series benchmark datasets in \textbf{Q2}, the Server Machine Dataset (SMD)  \cite{su2019robust}, and the NASA Telemetry Anomaly Detection (NASA) dataset \cite{hundman2018detecting} 
to see how PAC-Wrap affects the performance of time series anomaly detectors. The detailed result is reported in Appendix \ref{appendix:time}.


In \textbf{Q3}, we use the same experimental setup on the MNIST dataset \cite{deng2012mnist} as in \citep{lei2014classification}: 
we regard the digits $\{0,6,9\}$ as class "0" and digit $\{8\}$ as class "1". 
The training dataset contains 3044 images, with 541 in class 1. The test dataset contains 872 images, with 166 in class 1.
As in \citep{lei2014classification}, we train $\ell_1$-penalized logistic regression on two-thirds of the training data points and use the remaining one-third as the calibration data to identify the Conformal/PAC prediction sets. In the calibration dataset, we have 865 images in class 0 and 170 images in class 1.

Finally, we use the same synthetic dataset in \textbf{Q4} as in \textbf{Q1}.
\vspace{-5pt}
\subsubsection{Anomaly Detectors}
We consider the following anomaly detectors:
\begin{itemize}
\item \textbf{Isolation Forest} (IF) \cite{liu2012isolation} is an unsupervised model based on decision trees. 
\item \textbf{Local Outlier Factor} (LOF) \cite{breunig2000lof} is an unsupervised anomaly detection method which compares an estimated density of a data point to its neighbors.
\item \textbf{DevNet} \cite{Pang2019DeepAD} is a semi-supervised anomaly detector that uses a few labeled anomalies to separate the anomalies from normal data points. 
\item \textbf{LSTM-based anomaly detector} \cite{bhatnagar2021merlion, hundman2018detecting} is commonly used for time series data. For SMD, we wrap around a standard LSTM-encoder-decoder-based anomaly detector  \cite{bhatnagar2021merlion}. For the NASA data, we use the proposed LSTM-based anomaly detector in \cite{hundman2018detecting}.
\end{itemize}
 If an anomaly score threshold is not explicitly identified for the above anomaly detectors, we use a threshold that maximizes the F1 score, i.e., the harmonic mean of precision and recall. The F1 score is often used as an efficacy measure in the anomaly detection literature.
\subsubsection{Metrics}
\label{metrics}
Let TP, TN, FP, FN be the number of true positives, true
negatives, false positives, and false negatives, respectively.
We focus on the three most important error rates in anomaly detection: FNR = FN/(FN+TP) and FPR = FP/(FP+TN), ERR = (FN+FP)/(FN+TP+FP+TN). We compare FPR, FNR and ERR to a user-specified error constraint  (e.g., $\epsilon =  0.05$). 
We repeatedly run the experiments and check if the\textit{ error constraint violation rate}, defined as the fraction of times the error rate is above $\epsilon$, is lower than a user-specified confidence constraint (e.g., $\delta = 0.05$). 
To compare PAC-Wrap with a conformal prediction-based baseline, we use the definition of ambiguity from \citep{lei2014classification}, estimated as the fraction of data points falling into the ambiguity region in the test dataset:
\begin{equation}
    \textit{Ambiguity} = \frac{\sum_{(x,y) \in Z_{\text{test}}} \mathbbm{1}(d(x) \in \mathcal{U})}{|Z_\text{test}|}.
    \label{eq:ambiguity}
\end{equation}

\subsection{Q1. Empirical Validation}
\label{Q1}
We first empirically validate the theoretical guarantees of our false negative and false positive PAC prediction sets.
To study how anomaly detector performance affects our guarantees, we experiment with two kinds of anomaly detectors, the Local Outlier Factor (LOF) and the Isolation Forest (IF). 
Additionally, to study how calibration set size affects our guarantees, we experiment with 50\%, 75\%, and 100\% of the calibration set.
In these experiments, we set $\epsilon_{\text{fn}}=\epsilon_{\text{fp}}=0.05$ and $\delta_{\text{fn}} = \delta_{\text{fp}}=0.05$ as our constraints. Besides, 4000 independent Monte Carlo trials on both synthetic and benchmark datasets are performed. Out of these trials, we compute the empirical error constraint violation rate, which is the fraction of trials where the FPR or FNR is above 0.05. 

Note that the PAC guarantee assumes an infinite population, but we only have a finite dataset. To address this problem,
we propose the following method. First, we combine the calibration and test datasets to form a known finite population $\mathcal{D}$. We aim to validate the PAC guarantee over the known finite population $\mathcal{D}$, which is convenient since we can enumerate the population. Next, we train the LOF and IF on the training set. For each Monte Carlo trial, we then sample with replacement a new calibration set from the known finite population, of the same size and anomaly ratio as the original calibration set. We construct false positive and false negative PAC prediction sets on each newly sampled calibration set. Finally, we compute the FPR and FNR of the constructed PAC prediction sets over the known finite population $\mathcal{D}$.

We report, in Table \ref{tab: delta_synth} (on synthetic data) and Table \ref{tab: delta_real} (on the \textit{thyroid} dataset), a two-sided 95\% Clopper–Pearson interval for the error constraint violation rate. If the interval covers 0.05 (or falls below that), the empirical results are consistent with the error and confidence constraints being satisfied.
 In Table \ref{tab: delta_synth} and Table \ref{tab: delta_real}, the PAC guarantee is corroborated by the results on the synthetic and benchmark datasets since all the intervals fall below 0.05. The guarantee holds regardless of calibration set size and anomaly detectors. As a result, our results empirically validate Theorems \ref{th_fp} and \ref{th_fn}. Another observation is that the constraint violation rates on the benchmark dataset are much lower than 0.05, which means that the constructed PAC prediction sets on the benchmark dataset are conservative. We further discuss this observation in Appendix \ref{conservativeness}.

\begin{center}
    \begin{table}
    \begin{adjustbox}{width=0.4\textwidth}
\begin{tabular}{cccc}
\hline
Violation & Val Size & IF & LOF                 \\ \hline
\multirow{3}{*}{Pr(FPR\textgreater{}0.05)} & 50\%     & [0.034, 0.046] & [0.033, 0.045] \\
                                          & 75\%     & [0.024, 0.034] & [0.020, 0.030] \\
                                          & 100\%    & [0.031, 0.043] & [0.030, 0.042] \\
\multirow{3}{*}{Pr(FNR\textgreater{}0.05)} & 50\%     & [0.033, 0.045] & [0.034, 0.047]  \\
                                          & 75\%     & [0.032, 0.044] & [0.035, 0.048]  \\
                                          & 100\%    & [0.035, 0.047] & [0.032, 0.044] \\ \hline
\end{tabular}
\end{adjustbox}
\caption{The rate of error constraint violation on the synthetic data. }
\vspace{-20pt}
\label{tab: delta_synth}
\end{table}
\end{center}

\begin{center}
\begin{table}
\begin{adjustbox}{width=0.4\textwidth}
\begin{tabular}{ccll}
\hline  
Violation & \multicolumn{1}{c}{Val Size} & \multicolumn{1}{c}{IF} & LOF  \\ 
\hline
\multirow{3}{*}{Pr(FPR \textgreater{} 0.05)} & \multicolumn{1}{c}{50\%} & \multicolumn{1}{c}{[0.024, 0.035]} & [0.022, 0.033] \\
& \multicolumn{1}{c}{75\%}     & \multicolumn{1}{c}{[0.007, 0.013]} & [0.006, 0.012] \\
& 100\%  & [0.025, 0.036] & [0.026, 0.037]\\
\multicolumn{1}{l}{\multirow{3}{*}{Pr(FNR \textgreater{} 0.05)}} & 50\% & [0.006, 0.012] & [0.006, 0.012] \\
\multicolumn{1}{l}{} & 75\% & [0.010, 0.018] & [0.014, 0.022] \\
\multicolumn{1}{l}{}  & 100\% & [0.012, 0.020] & [0.016, 0.025] \\ \hline
\end{tabular}
\end{adjustbox}
\caption{The rate of error constraint violation on the benchmark dataset.}
\vspace{-25pt}
\label{tab: delta_real}
\end{table}
\end{center}

\vspace{-30pt}
\subsection{Q2. Wrapper Effect}
\label{error}
In this section, we conduct experiments to check how PAC-Wrap affects the error rates of the underlying anomaly detector. Specifically, we apply Algorithm \ref{Strategy} to remove the ambiguity region and check if the final FPR, FNR and ERR are bounded by the error constraint. For brevity, we omit repeatedly verifying the confidence constraint, which is already tested in \textbf{Q1}.
We report the following values:
\begin{itemize}
    \item \textbf{$\text{FNR}_{\text{or}}$, $\text{FPR}_{\text{or}}$}: the original FNR and FPR of the anomaly detector without our wrapper.
    \item \textbf{$\text{FNR}_{\text{tt}}$, $\text{FPR}_{\text{tt}}$}: the FNR and FPR of our wrapper using two thresholds $\tfn, \tfp$, given the initial error constraints.
    \item \textbf{$\text{FNR}_{\text{th}}$, $\text{FNP}_{\text{th}}$}: the FNR and FPR of our wrapper using one final threshold $\tau$, given the (possibly relaxed) error constraint.
    \item \textbf{ERR}: the final error rate of our wrapper, which is defined in \ref{metrics}. It is a weighted combination of $\text{FNR}_{\text{th}}$ and $\text{FPR}_{\text{th}}$.
    \item \textbf{$\epsilon$}: the final error level our wrapper can guarantee.
    
\end{itemize}

For i.i.d. data, we take DevNet \citep{Pang2019DeepAD} as the baseline anomaly detector. 
We first train DevNet on the \textit{campaign}, \textit{celeba}, and \textit{census} datasets respectively using default hyperparameters. Second, we wrap the trained model with the constructed false negative and false positive PAC prediction sets. Then, we find that DevNet may not perform well enough to simultaneously satisfy the user-specified errors constraints with the PAC prediction sets. Therefore, we use Algorithm \ref{Strategy} to adaptively relax the error constraint to enable DevNet to fulfill a reasonable guarantee.

\begin{figure}
    \centering
        \centering
        \includegraphics[width=0.4\textwidth]{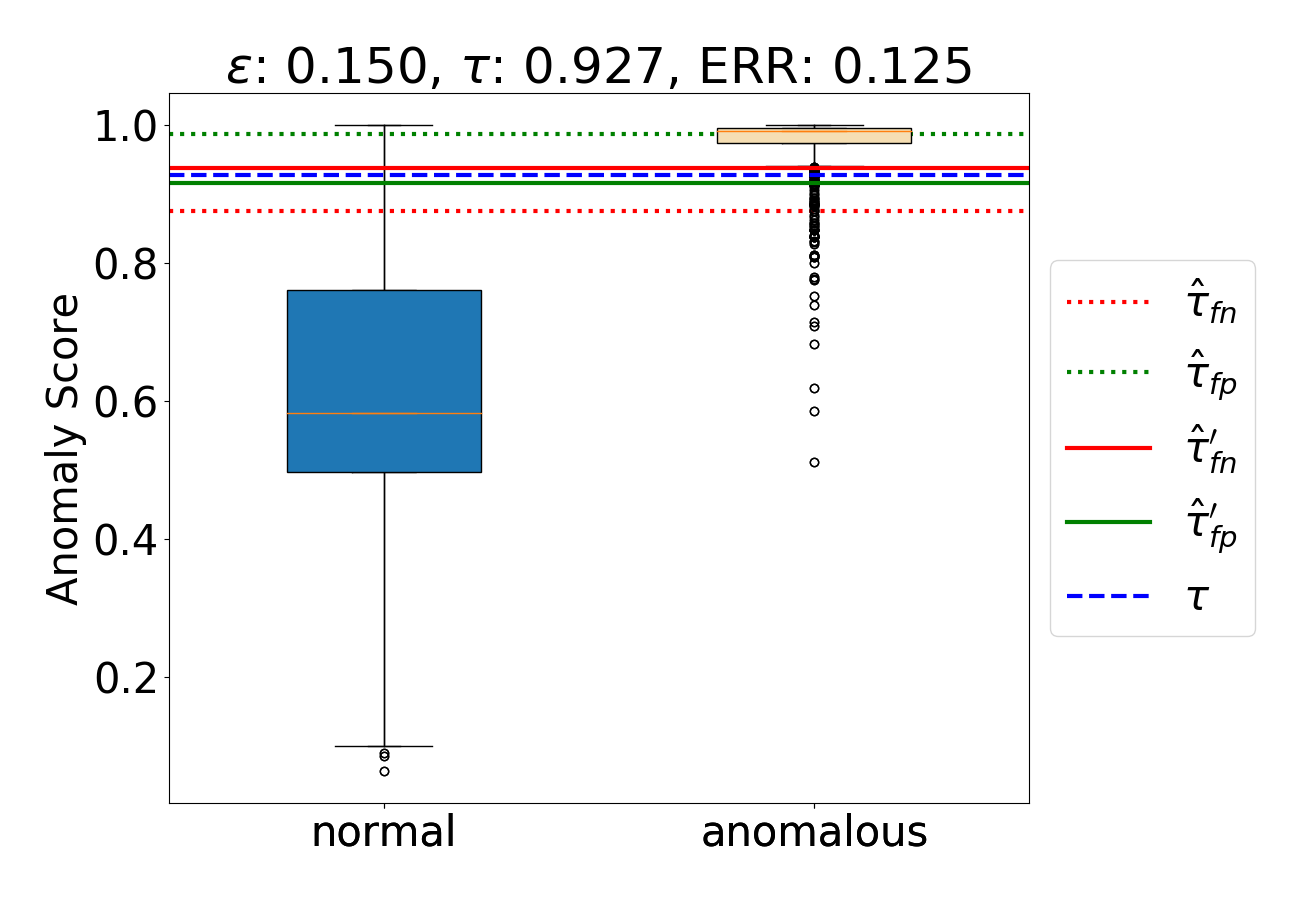}
        \vspace{-12pt}
        \caption{Box Plot and thresholds on the \textit{celeba} dataset using the DevNet anomaly detector. $\tfp > \tfn$ holds under the original error constraint ($\epsilon \leq 0.05$). By using Alg. \ref{Strategy}, $\tfn' \geq \tfp'$ under the relaxed error constraint ($\epsilon \leq 0.15$).}
            \vspace{-12pt}
    \label{benchmark}
\end{figure}

In Figure \ref{benchmark}, we show how this works on the \textit{celeba} dataset. 
Specifically, $\tfn$ and $\tfp$ are first chosen to satisfy the constraint $\epsilon_\fn = \epsilon_\fp=0.05, \delta_\fn=\delta_\fp=0.05$. 
Although most anomalies have higher anomaly scores than the normal data points, there is still considerable overlap. As a result, $\tfp$ is greater than $\tfn$ (the green dashed line is above the red dashed line in Figure \ref{benchmark}), which is an inconclusive case as discussed in \ref{subsec:adar}.
In other words, the anomaly detector cannot accurately distinguish the normal and the anomalous classes under the current error constraint, and we have to relax the constraint. 
After relaxing the error constraint by Algorithm \ref{Strategy}, we find $\epsilon=0.15$ and $\tfn' \geq \tfp'$ (solid red line is above solid green line in Figure \ref{benchmark}). Then, we can readily remove the ambiguity region by setting $\tau = (\tfn'+\tfp')/2$ (dashed blue line in Figure \ref{benchmark}) according to Theorem \ref{th_final}. 
A similar process occurs on the \textit{campaign} and \textit{census} datasets, resulting in the relaxed error constraints of $\epsilon=0.35$ and $\epsilon = 0.25$ respectively.

We report the detailed results for DevNet in Table \ref{tab: comp}. Table \ref{tab: comp} first shows that the $\text{FNR}_{\text{or}}$-s and $\text{FPR}_{\text{or}}$-s of DevNet violate the error constraint $\epsilon_\fn = \epsilon_\fp=0.05$ on \textit{campaign}, \textit{celeba}, and \textit{census} datasets. 
Then, columns $\text{FNR}_{\text{tt}}$ and $\text{FPR}_{\text{tt}}$ indicate that PAC-Wrap satisfies the original error constraints. 
After the constraint relaxation and ambiguity removal, columns $\text{FPR}_{\text{th}}$, $\text{FPR}_{\text{th}}$, and ERR are lower than the last column $\epsilon$, indicating that they all satisfy the relaxed error constraints. 
The underlying anomaly detectors determine the relaxed levels. Without our wrapper, the baselines can usually only control one of the FNR/FPR. 
Our method provides a principled way to balance the two rates and provides guarantees on their levels. 

\begin{center}
    \begin{table}
    \begin{adjustbox}{width=0.45\textwidth}
\begin{tabular}{lllllllll}
\hline
     & $\text{FNR}_{\text{or}}$ & $\text{FPR}_{\text{or}}$ & $\text{FNR}_{\text{tt}}$    & $\text{FPR}_{\text{tt}}$    & $\text{FNR}_{\text{th}}$ & $\text{FPR}_{\text{th}}$ & ERR & $\epsilon$ \\ \hline
campaign  & 0.000      & 0.998      & 0.026 & 0.043 & 0.267  & 0.259 & 0.266 & 0.35        \\
celeba & 0.029      & 0.456    & 0.026 & 0.042 & 0.121  & 0.097 & 0.120 & 0.15       \\ 
census & 0.055  & 0.561 & 0.048 & 0.047 & 0.230 & 0.202 & 0.229 & 0.25 \\
\hline
\end{tabular}
\end{adjustbox}
\caption{Error rate with PAC-Wrap wrapped around DevNet on i.i.d. data. Guarantees on FNR and FPR are met. After removing the ambiguity region, the $\text{FPR}_{\text{th}}$, $\text{FPR}_{\text{th}}$, and ERR satisfy the error constraints.}
\vspace{-20pt}
\label{tab: comp}
\end{table}
\end{center}

\vspace{-17pt}
Time series data are beyond the independence assumptions required for the PAC property, but can be transformed to reduce the dependence across time. We show the detailed result in the in Appendix \ref{appendix:time} that, in certain cases, our wrapper is also effective for time series anomaly detectors.  
Initially, we set $\epsilon_\fn = \epsilon_\fp=0.05, \delta_\fn=\delta_\fp=0.05$ and find that the sample size is occasionally too small to satisfy the error and confidence constraints. This is because, given a user-specified $\epsilon$ and $\delta$, we have a minimum requirement for the number of data points.
According to Theorem 1 in \cite{park2020pac}, the number of data points $n$ should  be at least $\log(1/\delta)/\log (1-\epsilon)$. For instance, if $\epsilon=\delta=0.05$, the minimum required sample sizes for labeled normal and anomalous data points are both 59.
When only limited labeled anomalies are available, we can relax the error and confidence level to give PAC guarantees. In the time series experiments where only 30 to 40 labeled anomalies are available, we set $\epsilon_\fn = \epsilon_\fp=0.10$,  $\delta_\fn=\delta_\fp=0.10$ to compute the thresholds.
After training the LSTM-based anomaly detectors on the training set, we find that they have the same performance issue as DevNet. Hence, we perform a similar constraint relaxation and ambiguity removal procedure.

\begin{figure}
    \centering
    \subfloat[The $\epsilon=0.4$ error constraint does not hold for the original anomaly detector.]{{\includegraphics[width=0.7\linewidth]{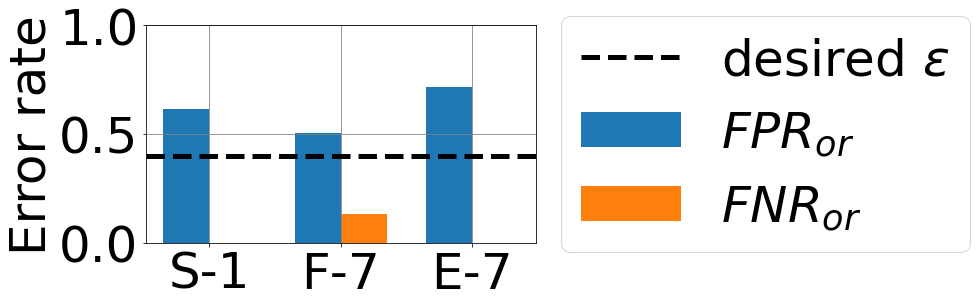} }}
    \qquad
    \subfloat[With our wrapper, the $\epsilon=0.4$ error constraint is met.]{{\includegraphics[width=0.7\linewidth]{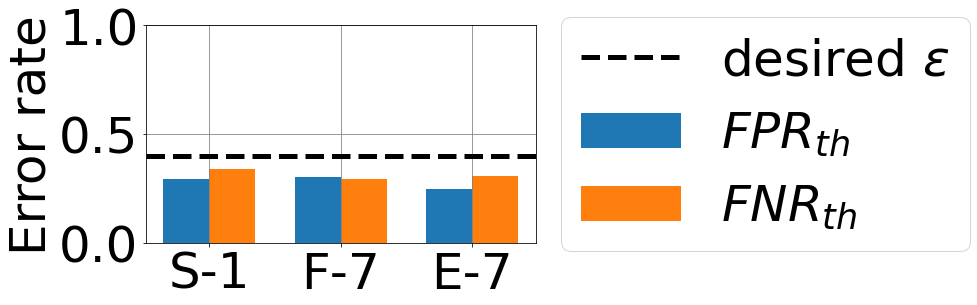} }}
    \vspace{-12pt}
    \caption{Results on the NASA data with $\epsilon = 0.4$. For the NASA anomaly detector, PAC-Wrap helps balance the FNR and FPR.}
    \label{fig: NASA}
    \vspace{-8pt}
\end{figure}

We show in Figure \ref{fig: NASA} some representative channels from the NASA dataset. The LSTM-based anomaly detector violates the error constraint, but PAC-Wrap controls both $\text{FPR}_{\text{th}}$ and $\text{FNR}_{\text{th}}$ to be smaller than $\epsilon$. 
For example, for the S-1, F-7, and E-7 channels, $\text{FPR}_{\text{or}}$ is even greater than the relaxed error constraint $\epsilon=0.4$. With a moderate increase in $\text{FNR}_{\text{th}}$, our wrapper can ensure both $\text{FPR}_{\text{th}}$ and $\text{FNR}_\text{th}$ are below $\epsilon=0.4$.

\vspace{-2pt}

\subsection{Q3. Baseline Comparison}

We compare PAC-Wrap to the method in \citep{lei2014classification}---denoted as \textit{CPAD}---which also provides guarantees on the FNR and FPR by calculating per-class thresholds. 
The error and confidence constraints are set as $\epsilon_\fn = \epsilon_\fp=0.05, \delta_\fn=\delta_\fp=0.05$.
We use the same known finite population method as in \ref{Q1} for evaluation. Specifically, after constructing the training, calibration and test datasets as in \citep{lei2014classification}, we construct the known finite population by combining the calibration dataset with the test dataset. Next, we perform 300 independent Monte Carlo trials and compute the error constraint violation rates on the known finite population. To identify the conformal/PAC prediction sets, we sample with replacement a new calibration dataset with the same size and anomaly ratio as the original calibration dataset in each trial. After constructing the prediction sets, we evaluate the FNR and FPR on the known finite population. 

The average FPR and FNR over the 300 Monte Carlo trials are reported in Table \ref{tab_cp_epsilon}. The result shows that the average FNR-s and FPR-s of CPAD and PAC-Wrap are basically below 0.05, therefore satisfying the error constraints. This finding is consistent with the class-conditional guarantees of PAC-Wrap and CPAD. To evaluate the satisfaction of the confidence constraint, we report a two-sided 95\% Clopper–Pearson interval for the error constraint violation rate in Table \ref{tab_cp_delta}. The result shows that CPAD's violation rates are approaching 50\%, while that of PAC-Wrap are at the desired level (below 0.05). That is because CPAD's guarantee holds marginally over the training dataset, which differs from the conditional guarantee of PAC-Wrap. It is possible that an insufficiently representative calibration set is drawn, and PAC-Wrap accounts for the scenario via introducing the confidence parameter $\delta$. 
In contrast, standard conformal prediction-based methods like CPAD do not consider the data representativeness and cannot provide a training-set conditional guarantee with high confidence. While satisfying the error constraint with much higher probability, PAC-Wrap induces slightly higher ambiguity than that of CPAD, as shown in Table \ref{tab_cp_epsilon}. However, the increment in ambiguity is mostly tolerable, especially in safety-critical applications where the violation of the error constraint might lead to a catastrophe.
\begin{table}[ht]
\centering
\begin{tabular}{llll}
\hline
Dataset          & FPR  & FNR  & \text{Ambiguity}  \\
\hline
Desired  & 0.050      & 0.050    & 0          \\
CPAD  & 0.049   & 0.051  &  0.222   \\
PAC-Wrap & 0.038 & 0.020  & 0.345   \\
\hline
\end{tabular}
\caption{Average FNR and FPR for CPAD and PAC-Wrap. On Average, both CPAD and PAC-Wrap satisfy the error constraint.}
\label{tab_cp_epsilon}
\end{table}
\vspace{-15pt}

\vspace{-20pt}
\begin{table}[ht]
\begin{tabular}{lll}
\hline
Method  & $\Pr(\text{FPR} > 0.05)$             & $\Pr(\text{FNR} > 0.05)$            \\ \hline
 CPAD    & [0.344, 0.458] & [0.498, 0.614] \\
PAC-Wrap  & [0.001, 0.024] & [0.000, 0.018] \\ \hline
\end{tabular}
\caption{Comparison of the error constraint violation rate of the CPAD and PAC-Wrap. CPAD violates the error constraint for nearly 50\% of the time and hence fails the confidence constraint. PAC-Wrap satisfies the 0.05 confidence constraint.}
\label{tab_cp_delta}
\end{table}

\vspace{-30pt}
\subsection{Q4. Ablation Study}
\begin{figure}
    \centering
    \includegraphics[scale=0.4]{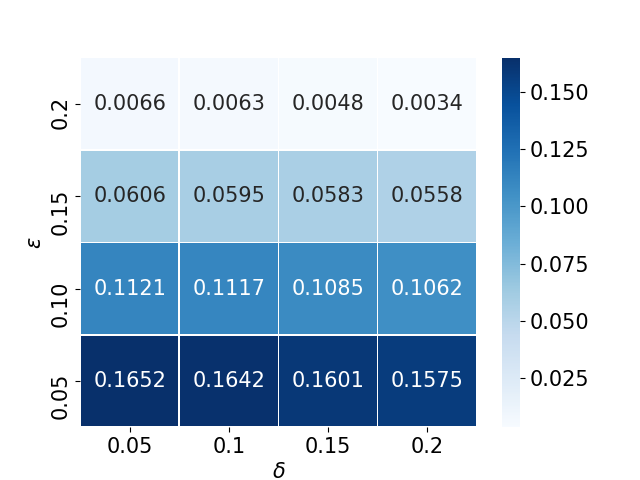}
    \vspace{-10pt}
    \caption{The average ambiguity as a function of $\epsilon$ and $\delta$. As $\epsilon$ and $\delta$ grow, the ambiguity shrinks.}
    \vspace{-15pt}
    \label{fig:trade-off}
\end{figure}

In this experiment, we want to see how the ambiguity (defined in Equation \eqref{eq:ambiguity}) changes with respect to the error parameter $\epsilon$ and confidence parameter $\delta$. Specifically, we set $\epsilon_\fn = \epsilon_\fp=\epsilon, \delta_\fn=\delta_\fp=\delta$. We then vary $\epsilon$ and $\delta$, and construct false positive and false negative PAC prediction sets on the synthetic dataset. For every combination of error parameter and confidence parameter, we do 100 Monte Carlo trials and compute the average ambiguity.
As shown in Figure \ref{fig:trade-off}, the ambiguity monotonically decreases with respect to $\epsilon$ and $\delta$. It suggests that there is an empirical trade-off between the constraints and ambiguity. We can relax constraints to decrease the ambiguity or vice versa. Moreover, $\epsilon$ has a larger effect on the ambiguity than $\delta$. 

\vspace{15pt}


\vspace{-20pt}
\section{Conclusion and Discussion}
We have developed a general framework called PAC-Wrap for guarantees in semi-supervised anomaly detection. 
Given many normal data points and a small number of anomalous data points, we use PAC-Wrap to control the false negative rate (FNR) and false positive rate (FPR). We conduct experiments on synthetic and benchmark datasets with various anomaly detectors to showcase the effectiveness of PAC-Wrap. Our method can readily wrap around virtually any existing anomaly detection algorithm, 
making our framework an off-the-shelf tool to provide rigorous PAC guarantees for these algorithms. 
Our method can be applied to safety-critical applications such as autonomous vehicles, surveillance video, and tumor diagnosis. By leveraging a limited number of labeled datapoints, PAC-Wrap can guarantee the FPR and FNR of anomaly detectors, which is highly important.

PAC-Wrap can be directly extended to a multi-class framework to provide conditional guarantees for each class for an immediate next step. 
One limitation of PAC-Wrap is that if the normal and anomalous distributions in the testing stage are significantly different from those in the calibration stage, the false negative and false positive guarantees might not hold, since PAC-Wrap cannot automatically adapt to the distribution shift. 
To see how distribution shifts affect the guarantees, we show an additional experiment in Appendix \ref{distribution shift}. 
In future work, it is important to enable PAC-Wrap to adapt to distribution shift during the testing stage.
\section{Acknowledgments}
This work was supported in part by DARPA/AFRL FA8750-18-C-0090, ARO W911NF-20-1-0080, ONR N00014-20-1-2744, NSF-1915398, NSF-2125561,
NSF-1934960, NSF-2046874
and SRC Task 2894.001. Any opinions, findings and conclusions or recommendations expressed in this material are those of the authors and do not necessarily reflect the views of the Air Force Research Laboratory (AFRL), the Army Research Office (ARO), the Defense Advanced Research Projects Agency (DARPA), the Office of Naval Research (ONR) or the Department of Defense, or the United States Government. 

\bibliographystyle{ACM-Reference-Format}
\bibliography{main}
\appendix
\section{Conservativeness}
\label{conservativeness}
The conservativeness over \textit{thyroid} dataset could be explained by the fact that the calibration set size (432) is significantly smaller than that of synthetic dataset (4000). Since a smaller calibration set is less representative of the true distribution, Equation \eqref{eqn:algorithm} will construct a possibly over-conservative prediction set to satisfy the confidence constraint, which leads to the violation rate being much lower than the confidence constraint. Moreover, a small calibration set, which is unrepresentative of the true distribution, could also contribute to a high violation rate. 
The fact that the calibration set is small could have the opposite effects on the violation rates. On the \textit{thyroid} dataset, the effect of Equation \eqref{eqn:algorithm} is dominant. As a result, the constructed PAC prediction sets are relatively conservative.

\section{Proof of Corollary \ref{th_cfp}}
\label{pf_cfp}
We replace the original prediction set $L_{\Ds}(C)$ with $L_{\Ds_{\text{nm}}}(\cfp)$, setting $\epsilon=\epsilon_{\text{fp}}, \delta=\delta_{\text{fp}}$, and construct the false positive PAC prediction set via solving \eqref{eqn:algorithm}. By Theorem \ref{thm:pred_set}, we have
$
    \Prob_{Z \sim \Ds^n_{\text{nm}}}[ L_{\Ds_{\text{nm}}}(\cfp) \le \epsilon_{\text{fp}} ] \ge 1 - \delta_{\text{fp}}. 
$
Therefore, $\cfp$ is $(\epsilon_{\text{fp}}, \delta_{\text{fp}})$-correct. $\square$

\section{Proof of Theorem \ref{th_fp}}
\label{pf_fp}
We have
\begin{equation*} \label{eq_th_fp}
\begin{split}
&\Pr(\hat y =1 \mid y=0) = \mathbb{E}_{x\mid y=0}[\mathbbm{1}(\hat y = 1)] \\
&= \mathbb{E}_{x\mid y=0}[\mathbbm{1}(0 \notin \cfp(x))]
= \mathbb{E}_{x \mid y=0}[\mathbbm{1}( y \notin C_{\hat\tau_{\fp}}(x)]\\
&= \mathbb{E}_{x \mid y=0}[\ell_{\fp}^{01}(x)] 
=  L_{D_{\text{nm}}}(\cfp).
\end{split}
\end{equation*} By Corollary \ref{th_cfp}, we have
\begin{equation*}
 \Prob_{Z \sim D_{\text{nm}}^n}[ L_{D_{\text{nm}}}(\cfp) \le \epsilon_\fp ] \ge 1 - \delta_{\fp}.   
\end{equation*} Since $\Pr(\hat y =1 \mid y=0)=L_{D_{\text{nm}}}(\cfp)$, we find
\begin{equation*}
    \Prob_{Z \sim D_{\text{nm}}^n}[ \Prob(\hat y =1 \mid y=0) \le \epsilon_\fp ] \ge 1 - \delta_{\fp}. \square 
\end{equation*}

\section{Proof of Corollary \ref{th_cfn}}
\label{pf_cfn}
We replace the original prediction set $L_{\Ds}(C)$ with $L_{\Ds_{\text{ano}}}(\cfn)$, setting $\epsilon=\epsilon_{\text{fn}}, \delta=\delta_{\text{fn}}$, and construct the false positive PAC prediction set via solving \eqref{eqn:algorithm}. By Theorem \ref{thm:pred_set}, we have
$\Prob_{Z \sim \Ds^n_{\text{ano}}}[ L_{\Ds_{\text{ano}}}(\cfn) \le \epsilon_{\text{fn}} ] \ge 1 - \delta_{\text{fn}}$. 
Therefore, $\cfn$ is $(\epsilon_{\text{fn}}, \delta_{\text{fn}})$-correct.  $\square$

\section{Proof of Theorem \ref{th_fn}}
\label{pf_fn}
We have
\begin{equation*} \label{eq_th_fn}
\begin{split}
&\Pr(\hat y =0 \mid y=1) = \mathbb{E}_{x\mid y=1}[\mathbbm{1}(\hat y = 0)] \\
&= \mathbb{E}_{x\mid y=1}[\mathbbm{1}(1 \notin \cfn(x))]
= \mathbb{E}_{x \mid y=1}[\mathbbm{1}( y \notin C_{\hat\tau_{\fn}}(x)]\\
&= \mathbb{E}_{x \mid y=1}[\ell_{\fn}^{01}(x)] 
=  L_{D_{\text{ano}}}(\cfn).
\end{split}
\end{equation*} By Corollary \ref{th_cfn}, we have
\begin{equation*}
 \Prob_{Z \sim D_{\text{ano}}^n}[ L_{D_{\text{ano}}}(\cfn) \le \epsilon_\fn ] \ge 1 - \delta_{\fn}.   
\end{equation*} Since $\Pr(\hat y =0 \mid y=1)=L_{D_{ano}}(\cfn)$, we find
\begin{equation*}
    \Prob_{Z \sim D_{\text{ano}}^n}[ \Prob(\hat y =0 \mid y=1) \le \epsilon_\fn ] \ge 1 - \delta_{\fn}. \square 
\end{equation*}

\section{Proof of Theorem \ref{th_cad}}
\label{pf_ad}
When $d(x) \geq \tfn$ and $d(x) \geq \tfp$, by Equation \eqref{y_hat}, $\hat y = 1$. In this case, the error rate $\epsilon_{\text{ad}}$ equals to the FPR. (The anomaly detector's prediction is correct when $y=1$.) By Theorem \ref{th_fp}, we have
\begin{equation*}
    \Prob_{Z \sim D_{\text{nm}}^n} \left[ \Prob(\hat y = 1 \mid y=0) \le \epsilon_\fp \right] \ge 1 - \delta_\fp.
\end{equation*} In other words, the error rate when $d(x) \geq \tfn$ and $d(x) \geq \tfp$ satisfies 
\begin{equation*}
    \Prob_{Z_{\text{nm}} \sim D_{\text{nm}}^n} \left[ \epsilon_{\text{ad}} \le \epsilon_\fp \right] \ge 1 - \delta_\fp.
\end{equation*}

Similarly, when $d(x) \leq \tfn$ and $d(x) \leq \tfp$, by Equation \ref{y_hat}, $\hat y = 0$. In this case, the error rate $\epsilon_{\text{ad}}$ equals to the FNR. (The anomaly detector's prediction is correct when $y=0$.) By Theorem \ref{th_fn}, we have
\begin{equation*}
    \Prob_{Z \sim D_{\text{ano}}^n} \left[
    \Prob(\hat y = 0 \mid y=1)\le \epsilon_\fn \right] \ge 1 - \delta_\fn.
\end{equation*} Thus, the error rate when $f(x) \geq \tfn$ and $f(x) \geq \tfp$ satisfies 
\begin{equation*}
    \Prob_{Z_{\text{ano}} \sim D_{\text{ano}}^n} \left[ \epsilon_{\text{ad}}  \le \epsilon_\fn \right] \ge 1 - \delta_\fn.
\end{equation*}

Therefore, if we make a certain prediction by Equation \eqref{y_hat},
we can bound the error rate as
\begin{equation*}
    \begin{split}
        \epsilon_{\text{ad}} &= \epsilon_{\text{fp}} \cdot \Pr(y=0) + \epsilon_{\text{fn}} \cdot \Pr(y=1)\\
        &\leq \max{(\epsilon_{\text{fp}}, \epsilon_{\text{fn}})} \cdot \Pr(y=0) + \max{(\epsilon_{\text{fp}}, \epsilon_{\text{fn}})} \cdot \Pr(y=1)\\
        &= \max{(\epsilon_{\text{fp}}, \epsilon_{\text{fn}})}.
    \end{split}
\end{equation*}
The inequality holds with probability at least  $1-(\delta_{\text{fp}} + \delta_{\text{fn}})$ due to the union bound. Thus, the claim follows. $\square$

\section{Proof of Theorem \ref{th_final}}
\label{pf_th_margin}


Since Algorithm \ref{Strategy} returns $\tfp', \tfn'$ and $\epsilon$ only when $ \tfn' \geq \tfp'$, we first prove that the error rate of the anomaly detector is bounded by $\epsilon$.
Following the proof for Theorem \ref{th_cad}, we have  the guarantee that the FNR and FPR of the anomaly detector is bounded by the updated $\epsilon_{\text{fn}}$ and $\epsilon_{\text{fp}}$ in Algorithm \ref{Strategy}, using $\tfn'$ and $\tfp'$ respectively. 
If we use a threshold $\tau' < \tfn'$, the corresponding FNR, denoted as $\epsilon'$, obeys
\begin{equation}
    \epsilon' \leq \epsilon_{\text{fn}} \leq \epsilon.
    \label{eq: th_final_fn}
\end{equation}
This is because using a lower threshold corresponds to a lower quantile of the lower tail part for $\hat y = 1$ distribution, and we have a smaller chance of making false negative prediction, i.e., classifying an anomaly as a normal point.

Similarly, if $\tau' > \tfp'$, for the FPR, denoted as $\epsilon'$, we will have:
\begin{equation}
    \epsilon' \leq \epsilon_{\text{fp}} \leq \epsilon.
    \label{eq: th_final_fp}
\end{equation}
The same logic follows here; a higher threshold corresponds to a higher quantile of the upper tail of the distribution $x|\hat y = 0$. Hence we have a smaller chance of making false positive prediction.

Since $\tfp \leq \tau \leq \tfn$, let $\tau' = \tau$. Based on Theorem \ref{th_cad}, equation (\ref{eq: th_final_fn}) and equation (\ref{eq: th_final_fp}), the error rate of the anomaly detector $\epsilon_\text{ad}$ is bounded by $\epsilon$:
\begin{equation*}
\begin{split}
   \epsilon_{\text{ad}} &= \max{(\epsilon_{\text{fp}}, \epsilon_{\text{fn}})}
   \leq \max{(\epsilon, \epsilon)}
   = \epsilon.
\end{split}
\end{equation*} 
Since we use $\delta_\text{fn}, \delta_\text{fp}$ to re-calculate $\tfp', \tfn'$, the resulting $\delta$ can be taken as $\delta_\text{fn}+\delta_\text{fp}$ according to Theorem \ref{th_cad}. Therefore, the claim follows. $\square$

\section{Proof of Lemma \ref{ambiguity}}
\label{pf_lma}

We first prove that if $\tfn \geq \tfp$, then $\sum_{(x,y) \in Z_{\text{nm}}} \mathbbm{1}(d(x) > \tfn) \leq k_\fp^*$ and $ \sum_{(x,y) \in Z_{\text{ano}}} \mathbbm{1}(d(x) < \tfp) \leq k_\fn^*$. 
To see this, we construct a false positive PAC prediction set using \eqref{eqn:algorithm} and make a prediction using \eqref{fn_construct}. Therefore, we have
$$
 \sum_{(x,y) \in Z_{\text{nm}}} \mathbbm{1}(d(x) > \tfp) \leq k_\fp^*.
$$
Since $\tfn > \tfp$, we find
\begin{equation*}
\begin{split}
    \sum_{(x,y) \in Z_{\text{nm}}} \mathbbm{1}(d(x) > \tfn) &\leq \sum_{(x,y) \in Z_{\text{nm}}} \mathbbm{1}(d(x) > \tfp) \leq  k_\fp^*.
\end{split}
\end{equation*}

Similarly, for the false negative PAC prediction set, we have
\begin{equation*}
\begin{split}
     \sum_{(x,y) \in Z_{\text{ano}}} \mathbbm{1}(d(x) < \tfp) &\leq \sum_{(x,y) \in Z_{\text{ano}}} \mathbbm{1}(d(x) < \tfn) 
     \leq k_\fn^*. 
\end{split}
\end{equation*}

Next, we prove that if $\sum_{(x,y) \in Z_{\text{nm}}} \mathbbm{1}(d(x) > \tfn) < k_\fp^*$ and $\sum_{(x,y) \in Z_{\text{ano}}} \mathbbm{1}(d(x) < \tfp) < k_\fn^*$, then $\tfn \geq \tfp$. We argue by contradiction. 
Suppose that $\sum_{(x,y) \in Z_{\text{nm}}} \mathbbm{1}(d(x) > \tfn) < k_\fp^*$, and $\sum_{(x,y) \in Z_{\text{ano}}} \mathbbm{1}(d(x) < \tfp) < k_\fn^*$, but $\tfp < \tfn$. Then, for the false negative PAC prediction set, we should choose $\tfp$ instead of $\tfn$, since the identified $\tfn$ should be the largest threshold satisfying $\hat \epsilon_\fn$ and $\tfp > \tfn$. This contradicts that $\tfn$ is the chosen threshold. As a result, our assumption does not hold, and we have $\tfn \geq \tfp$. 

In summary, $\tfn \geq \tfp$ if and only if $\sum_{(x,y) \in Z_{\text{nm}}} \mathbbm{1}(d(x) > \tfn) < k_\fp^*$ and $\sum_{(x,y) \in Z_{\text{ano}}} \mathbbm{1}(d(x) < \tfp) < k_\fn^*$. $\square$


\section{Time-series Experiments}
\label{appendix:time}
We also experiment with two challenging time series anomaly detection datasets,  the Server Machine Dataset  \cite{su2019robust}, and NASA Telemetry Anomaly Detection \cite{hundman2018detecting}, to illustrate the effectiveness of PAC-Wrap on sequential data.
The NASA dataset consists of spacecraft telemetry data like radiation, temperature, and power from the Soil Moisture Active Passive satellite (SMAP), and the Curiosity Rover on Mars (MSL). In addition, it contains 193500 records for training and 501346 records for testing, of which around 10\% are anomalies.  
SMD is a dataset collected from a large Internet company over five weeks, with 38 features such as CPU load, network usage,
and memory usage. It contains a training set of 708405 records and a test set of 708420 records, among them 4.16\% are anomalies. We split the original test set into a calibration set (20\%) and a final test set (80\%) for both SMD and NASA.
\begin{center}
    \begin{table}
    \begin{adjustbox}{width=0.5\textwidth}
\begin{tabular}{llllllllll}
\hline
     & $\text{FNR}_{\text{or}}$ & $\text{FPR}_{\text{or}}$ & $\text{FNR}_{\text{tt}}$    & $\text{FPR}_{\text{tt}}$     & $\text{FNR}_{\text{th}}$ & $\text{FPR}_{\text{th}}$ & ERR & $\epsilon$ \\ \hline
S-1 & 0.000      & 0.615      & 0.059 & 0.088 & 0.340  & 0.293  & 0.337 & 0.40    \\
F-7 & 0.132      & 0.503      & 0.060 & 0.071 & 0.292  & 0.304  & 0.293 & 0.40    \\
E-7 & 0.000     & 0.714      & 0.076 & 0.081 & 0.306  & 0.246  & 0.304 & 0.40     \\
T-1 & 0.001     & 0.653      & 0.103 & 0.099 & 0.367  & 0.448  & 0.382 & 0.50     \\
T-2 & 0.011      & 0.738      & 0.063 & 0.084 & 0.384  & 0.428  & 0.393 & 0.50     \\
P-3 & 0.013      & 0.724      & 0.053 & 0.065 & 0.363  & 0.379  & 0.366 & 0.50  \\
\hline
\end{tabular}
\end{adjustbox}
\caption{Error rate with PAC-Wrap applied to the LSTM-based anomaly detector on the NASA data. First column is the corresponding channels. $\text{FNR}_{\text{tt}}$ and $\text{FPR}_{\text{tt}}$ satisfy the $\epsilon = 0.1$ guarantees. After removing the ambiguity region, the $\text{FNR}_{\text{th}}$, $\text{FPR}_{\text{th}}$, and ERR satisfy the relaxed error constraints.}
\vspace{-25pt}
\label{tab: comp_ts}
\end{table}
\end{center}



\begin{table}
\begin{adjustbox}{width=0.5\textwidth}
\begin{tabular}{lllllllll}
\hline
   & $\text{FNR}_{\text{or}}$ & $\text{FPR}_{\text{or}}$ & $\text{FNR}_{\text{tt}}$    & $\text{FPR}_{\text{tt}}$     & $\text{FNR}_{\text{th}}$ & $\text{FPR}_{\text{th}}$ & ERR &$\epsilon$  \\ \hline
15-60  & 0.036      & 0.937      & 0.074 & 0.087 & 0.487  & 0.474  & 0.487 & 0.60     \\
15-120 & 0.368      & 0.563      & 0.052 & 0.000 & 0.598  & 0.359  & 0.587 & 0.60     \\
15-240 & 0.011      & 0.918      & 0.086 & 0.088 & 0.538  & 0.399  & 0.532 & 0.60     \\
30-60  & 0.077      & 1.000      & 0.077 & 0.000 & 0.780  & 0.201  & 0.748 & 0.80     \\
30-120 & 0.818      & 0.127      & 0.057 & 0.000 & 0.781  & 0.174  & 0.746 & 0.80     \\
30-240 & 0.379      & 0.467      & 0.058 & 0.000 & 0.738  & 0.158  & 0.706 & 0.80     \\
\hline
\end{tabular}
\end{adjustbox}
\caption{Error rate with PAC-Wrap wrapped around the LSTM-based anomaly detector on the SMD data. First column is the corresponding combinations. $\text{FNR}_{\text{tt}}$ and $\text{FPR}_{\text{tt}}$ satisfy the $\epsilon = 0.1$ guarantees. After removing the ambiguity region, the $\text{FNR}_{\text{th}}$, $\text{FPR}_{\text{th}}$, and ERR satisfy the relaxed error constraints.}
\vspace{-25pt}
\label{tab: comp_smd}
\end{table}
 The detailed result for the NASA data is reported in Table \ref{tab: comp_ts}.
In the T-1, T-2, and P-3 channels, both $\text{FPR}_{\text{th}}$ and $\text{FNR}_\text{th}$ are guaranteed to be smaller than the relaxed error constraint, and the final error rate ERR is also below the required error constraint $\epsilon = 0.5$. 
Wrapped around the original NASA anomaly detector, PAC-Wrap can reduce the gap between the FNR and FPR and thus has a more balanced performance. 

Since there are more datapoints in the SMD dataset than in the NASA one, to approach the independence condition formally required by our guarantees, 
we consider the windows of the first 15 and 30 contiguous timesteps as data points for every 60, 120, and 240 timesteps. As shown in Table \ref{tab: comp_smd}, $\epsilon=0.6, 0.8$ is the relaxed error constraint given the anomaly score distribution. 
For the baseline anomaly detector, $\text{FPR}_{\text{or}}$-s sometimes fail the $\epsilon=0.6$ guarantee for the 15-timestep settings. 
For all the 30-timestep settings, the original anomaly detectors violate the $\epsilon=0.8$ guarantee on either $\text{FPR}_{\text{or}}$ or $\text{FNR}_{\text{or}}$. 
However, using PAC-Wrap as a wrapper, we ensure that both $\text{FNR}_{\text{th}}$ and $\text{FPR}_{\text{th}}$ fall below 0.6 and 0.8. 
The final error rates (ERR) are smaller than the maximum of $\text{FNR}_{\text{th}}$ and $\text{FPR}_{\text{th}}$, which also empirically supports Theorem \ref{th_cad}.

\vspace{-5pt}
\section{Distribution shift}
\label{distribution shift}
\begin{figure}[H]
    \centering
    \includegraphics[scale=0.2]{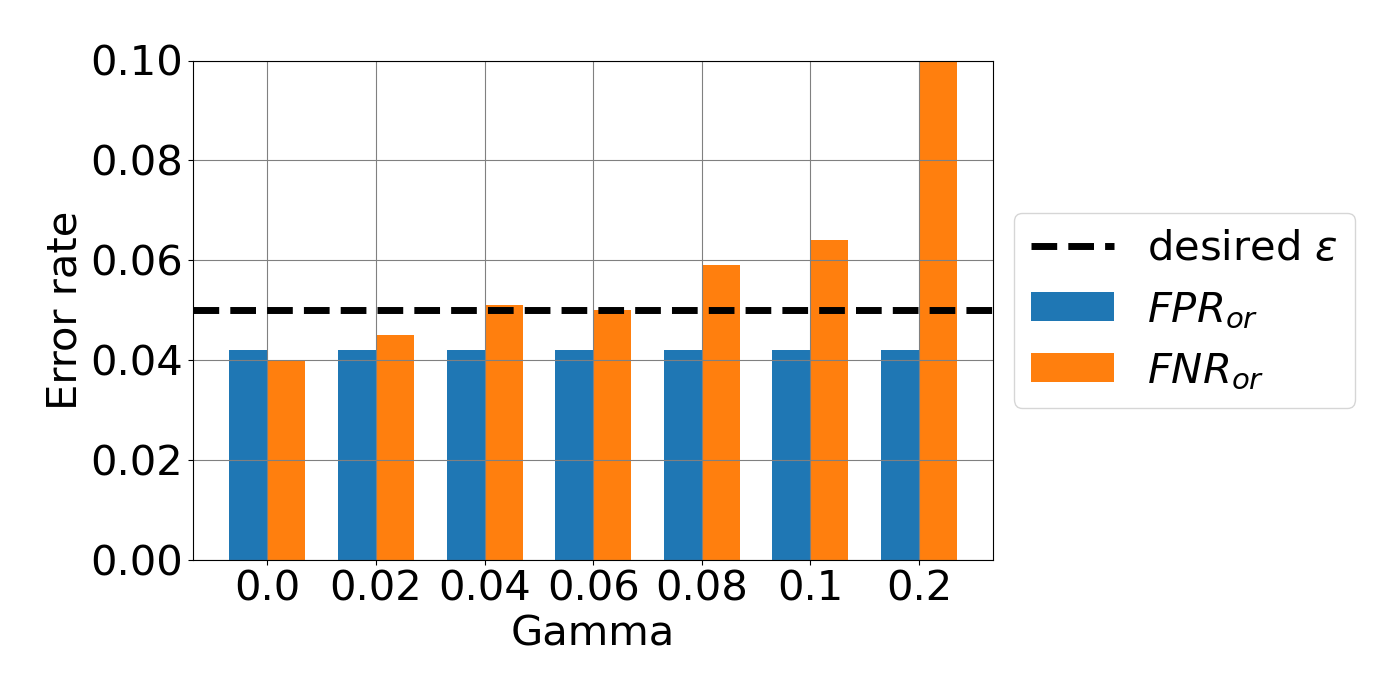}
    \caption{FPR and FNR after mixing different anomaly distributions}
    \label{fig:shift}
\end{figure}

To see how shifts in the anomaly distribution affect our guarantees, we generate data from three distributions in the following way:
\begin{align*}
    X_{\text{normal}} &\sim \mathcal{N}(\mu_{\text{normal}}, \sigma^2 I_p) \\
    X_{\text{anomalous}} &\sim \mathcal{N}(\mu_{\text{anomalous}}, \sigma^2 I_p)\\
    X_{\text{mixture}} &\sim \mathcal{N}(\gamma \cdot \mu_{\text{normal}} + (1-\gamma) \cdot\mu_{\text{anomalous}}, \sigma^2 I_p),
\end{align*} where $\gamma \in [0,1]$ is a mixing ratio. We set $\mu_{\text{normal}}=[0, 0, 0, 0, 0]^\top$, $\mu_{\text{anomalous}} = [3, 3, 3, 3, 3]^\top$, and $\sigma=2.0$. 
Then, we construct the training set by sampling 98,000 data points from $\mathcal{N}(\mu_{\text{normal}}, \sigma^2 I_p)$; we construct the calibration set by sampling 1,000 anomalies from $\mathcal{N}(\mu_{\text{anomalous}}, \sigma^2 I_p)$; we construct the testing set by sampling 1,000 data points from $\mathcal{N}(\gamma \cdot \mu_{\text{normal}} + (1-\gamma) \cdot\mu_{\text{anomalous}}, \sigma^2 I_p)$ with different $\gamma$-s. We set $\gamma$ to $\{0, 0.02, 0.04$, $0.06, 0.08$, $0.1, 0.2\}$ and $\epsilon=\delta=0.05$. We run PAC-Wrap on the training, calibration, and testing sets. As shown in Figure \ref{fig:shift}, guarantees nearly hold when $\gamma$ equals to 0.02, 0.04, and 0.06. However, when the mixing rate is too large, the guarantees might fail to hold.

\end{document}